\documentclass[10pt,twocolumn,letterpaper]{article}

\usepackage{ijcb}
\usepackage{times}
\usepackage{epsfig}
\usepackage{graphicx}
\usepackage{amsmath}
\usepackage{amssymb}
\usepackage[affil-it]{authblk}
\usepackage{makecell}
\usepackage{url}
\usepackage{hyperref}
\usepackage{tablefootnote}

\ijcbfinalcopy 

\ifijcbfinal\fi
\begin{document}

\title{DFGC 2022: The Second DeepFake Game Competition}


\author{
Bo Peng\textsuperscript{1}, Wei Xiang\textsuperscript{1}, Yue Jiang\textsuperscript{1}, Wei Wang\textsuperscript{1}, Jing Dong\textsuperscript{1}\thanks{Jing Dong is the corresponding author. 

This work is supported by 
the National Natural Science Foundation of China (NSFC) under Grants 61902400, U19B2038, 61972395 and a grant from Young Elite Scientists Sponsorship Program by CAST (YESS).}, Zhenan Sun\textsuperscript{1}, Zhen Lei\textsuperscript{2}, Siwei Lyu\textsuperscript{3} 
}
\affil{
\textsuperscript{1}Center for Research on Intelligent Perception and Computing (CRIPAC), Institute of Automation, Chinese Academy of Sciences, Beijing 100190, China \\
\textsuperscript{2}National Laboratory of Pattern Recognition (NLPR), Institute of Automation, Chinese Academy of Sciences, Beijing 100190, China \\
\textsuperscript{3}University at Buffalo, State University of New York, NY14260, USA 
}

\maketitle
\thispagestyle{empty}

\begin{abstract}
   This paper presents the summary report on our DFGC 2022 competition.
   The DeepFake is rapidly evolving, and realistic face-swaps are becoming more deceptive and difficult to detect. On the other hand, methods for detecting DeepFakes are also improving. There is a two-party game between DeepFake creators and defenders. This competition provides a common platform for benchmarking the game between the current state-of-the-arts in DeepFake creation and detection methods. The main research question to be answered by this competition is the current state of the two adversaries when competed with each other. This is the second edition after the last year's DFGC 2021, with a new, more diverse video dataset, a more realistic game setting, and more reasonable evaluation metrics. With this competition, we aim to stimulate research ideas for building better defenses against the DeepFake threats. We also release our DFGC 2022 dataset contributed by both our participants and ourselves to enrich the DeepFake data resources for the research community (\url{https://github.com/NiCE-X/DFGC-2022}).
\end{abstract}

\section{Introduction}
After the DeepFake first broke out in 2017 \cite{url_deepfake}, it has drawn lots of concerns from the public for its potential harms to information credibility and democracy. The word ``DeepFake" now refers to a variety of AI-enabled high-quality synthesis contents depicting altered and generated imagery of human subjects, e.g. face swap, face reenactment, lip-syncing, GAN generation, facial attribute editing. 
Face swap is a well-known and very popular form of Deepfake.
Open-source face swap tools are flourishing with notable projects like DeepFaceLab \cite{url_DeepFaceLab}, deepfakes/faceswap \cite{url_faceswap}.
Given its popularity and high quality, in this competition, we focus on the face swap form of DeepFake.

This competition features the evaluation of both DeepFake creation and DeepFake detection in an adversarial and dynamic game manner. Most previous competitions only focus on evaluating DeepFake detection methods based on some fixed DeepFake datasets, e.g., the Facebook DFDC \cite{DFDC}, the ForgeryNet Challenge \cite{ForgeryNet} and the FaceForensics++ Benchmark \cite{FaceForensics++}. Recent DeepFake competitions incorporate larger and more diverse datasets and also more tasks, e.g. image segmentation and video localization.
However, there have not been any other competitions that evaluate both DeepFake detection and creation in an adversary manner to the best of our knowledge. 

This is the 2nd edition of the DeepFake Game Competition after its was first held in the last year \cite{DFGC2021}. This edition inherits the general idea and workflow from its precedent, but it features several new improvements. Compared with DFGC 2021, the dataset was composed of videos instead of images to better imitate real-world situations, a new consented dataset was collected to be used as materials for creating the DeepFake, the evaluation additionally included some subjective metrics for better completeness, and some minor defects from the last year were mended, e.g. previously used SSIM and objective noise score cannot faithfully reflect the subjective quality assessment in some cases. 

\section{Related Work}
In this section, we discuss representative state-of-the-arts methods in both face-swap creation and DeepFake detection. This will help link top solutions in our competition to a  broader background in this field.
\subsection{Face Swap Methods}
The DeepFake model proposed by the Reddit user ``Deepfakes” is an encoder-decoder based face swap model. It incorporates a shared encoder for both the source face and the target face and two separate decoders for each one. In our vocabulary, the source face provides the facial ID while the target face provides the background and head periphery. 
Recent popular open-source face swap tool-kits \cite{url_DeepFaceLab, url_faceswap} develop on the encoder-decoder based method, and they improve the realism and quality with more sophisticated segmentation, blending and post-processing techniques. This line of methods produce very high-quality results, as usually seen on video platforms like YouTube. However, they have efficiency drawbacks as they require training separate models for each pair of faces to be swapped.

Another category of face-swap methods learn disentangled representations for the facial ID information and the attribute information and then fuse them from different images and input to a decoder to achieve the swap. A representative method is the FaceShifter model \cite{FaceShifter}, which is later improved by the more efficient SimSwap model \cite{SimSwap}, the InfoSwap model \cite{InfoSwap} with information bottleneck, and the FaceSwapper model \cite{FaceSwapper} that incorporates semantic segmentation information, etc.. These methods achieve generalization for the swap of arbitrary faces. To tackle face-swap at higher resolution, the MegaFS model \cite{MegaFS} employs the pre-trained StyleGAN \cite{StyleGAN} and learn the mapping in the latent space of StyleGAN. A unified framework for the face-swap and the expression reenactment tasks is proposed in \cite{unifiedSwap}, which employs the 3D Morphable Model for decomposition of unification.
\subsection{DeepFake Detection Methods}
Current detection methods can be broadly grouped into two categories, the clue-based ones and the learning-based ones. The first category identifies specific clues that indicate the fake nature. Representative approaches include using the defined visual artifacts \cite{visual_artifacts}, the lack of eye-blinking \cite{blinking}, the behavioral pattern of a specific person \cite{world_leader}, and the phoneme-viseme mismatch \cite{lip_match}, etc.. These clue-based methods are more interpretable, but they are less generalizable and can be by-passed by new generations of DeepFakes that repaired these clues. 

The other category of methods, i.e. the learning-based ones, directly learn discriminative features from real and fake data, and they are more actively studied in recent works and broadly used in competitions. According to the DFDC grand challenge report \cite{DFDC}, top solutions all use deep models, design various data-augmentation approaches and use model ensemble for more accurate and stable performance. For example, the 1st place model \cite{url_DFDC1} introduced comprehensive facial part erasing as strong data augmentation and ensembled seven EfficientNet-B7 models, the 2nd place model \cite{url_DFDC2} used attention-cropping and attention-dropping \cite{see_better} as data augmentations and ensembled one Efficient-B3 model and one Xception model. According to the more recent ForgeryNet challenge report \cite{ForgeryNet}, the 1st place model used global patch-wise consistency \cite{patch_consistency} and introduced a dynamic feature queue to help solve the catastrophic forgetting and fully mine hard samples. In our previous DFGC-2021 competition \cite{DFGC2021}, since unknown DeepFake creation methods and adversarial noises were used in the test data, top detection solutions used data augmentation approaches, such as generating new face-swaps on-the-fly and using adversarially attacked data in training. 

In recent literature, some work focus on designing effective attention-based modules to mine richer and more effective features \cite{On_the_detection, Multi-attention, representative_mining}. Self-supervised face-swap generation, as in Face X-ray \cite{face-Xray} and \cite{self_blend}, is a popular approach to easily augment the size and diversity of the training data. New losses such as single-center loss \cite{center_loss} and meta-learning inspired strategy \cite{meta_learn} are also investigated to achieve better generalization.

\section{The Overall Workflow}
\begin{figure*}[ht]
\centering
\centerline{\includegraphics[width=0.7\textwidth]{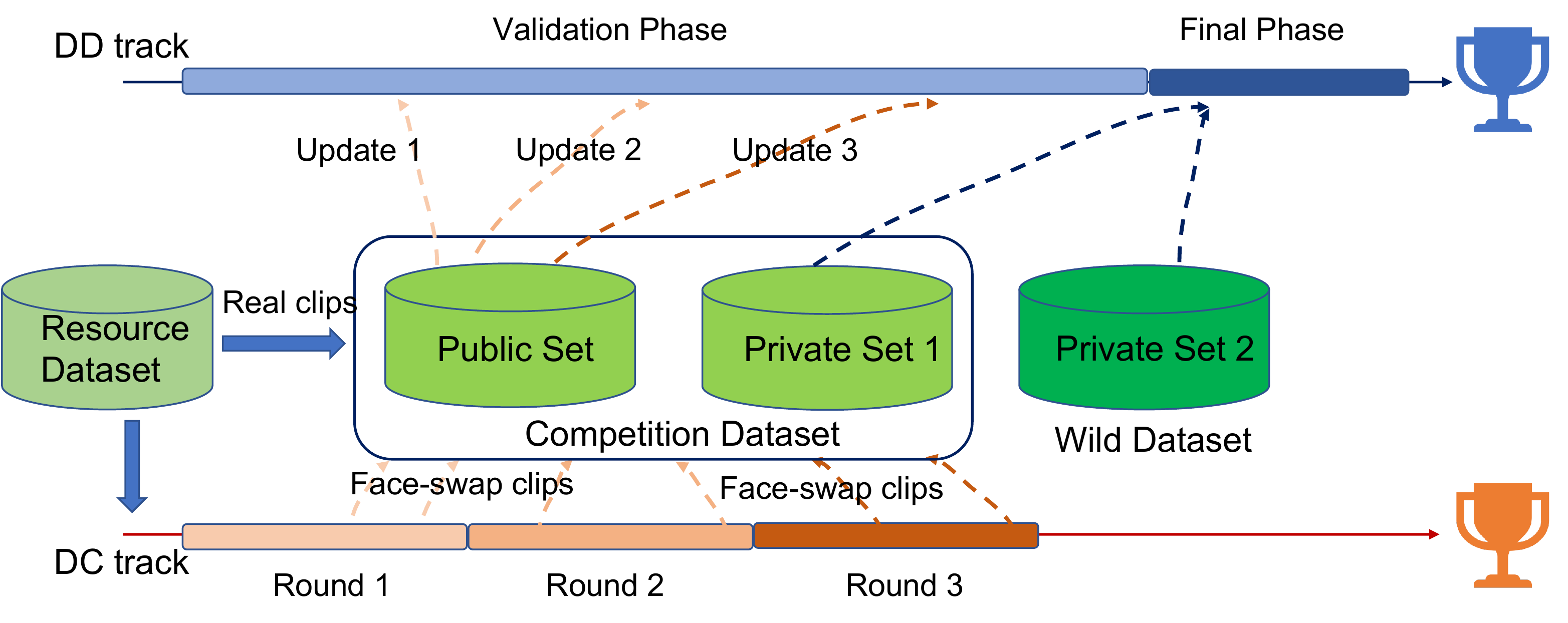}}
\caption{The overall workflow of DFGC 2022. It has two tracks: the DeepFake Detection (DD) track and the DeepFake Creation (DC) track. See the text for detailed descriptions.}
\label{fig_design}
\end{figure*}

The overall workflow of DFGC 2022 is shown in Fig. \ref{fig_design}. Our competition has two tracks: the DeepFake Creation (DC) track and the DeepFake Detection (DD) track. The DC track is composed of three submission rounds, and the DD track is composed of the validation phase and the final phase.

We first collected a diverse video dataset, which is named as the resource dataset, and provided it to the creation track participants to be used for creating face swap video clips. During the three submission rounds of the creation track, participants submitted their creation results for evaluation, and each team can submit up to two submissions, each containing 80 specified face-swap clips, for evaluation in each DC round. After the end of each DC round, the organizers evaluated the creation results with multiple subjective and objective metrics. In the meantime, submitted face-swap clips were combined with a set of real clips extracted from the resource dataset, they were post-processed with video compression operations, and finally partitioned into the public test set and the private test set with disjoint identities. These datasets were used as testing data in the DD track. We conducted compression to suppress the forgery artifacts and to make the detection more challenging.

In the DD track, the public test set was used in the validation phase, and the private set in the final phase. The public test set was released to the DD participants without groundtruth labels for detection prediction, and their prediction results were submitted back to the competition platform for evaluation in the validation phase. Each team can submit up to two submissions each day in the validation phase and see the AUC-ROC (the Area Under Curve of the Receiver Operating Characteristic curve) of their submissions on the leaderboard immediately. The public test set was dynamically enlarged with new data for three times after each of the three DC rounds. 
The private test set was retained by the organizers for the final phase evaluation, during which the top-10 teams in the validation phase should submit their inference codes to the organizers for evaluation on the private set. After checking the training codes for compliance with the competition rules, top-3 teams were determined by the AUC scores on the private test set.

For fairness, some rules on the detection track were made and declared in the competition. The detection track participants cannot use the released competition dataset for training or developing, but they can use any other publicly available datasets to train their models. Self-created face-swap data during off-line or on-line training (as data augmentation) is not allowed, unless the created data was made publicly available and announced during the competition. Remembering facial IDs or other shortcuts are not allowed, as the competition focus on general-purposed detection methods. 

As mentioned, we checked the top-3 training codes and data to ensure these rules were not violated. Their training codes were first sent to the organizers, and we inspected the codes carefully. We then connected with the top teams through video conference and inspected their training datasets and training process. They were then instructed to retrain the models three times, and we tested their retrained models on the public test set to compare with their scores on the leaderboard. Because of the randomness in training and the single-best-selection mechanism of the leaderboard, it is not 
surprising to see retrained models perform a little worse than the submitted one on the public set. The best retrained AUC score was within the $-0.02$ range of the leaderboard score, for all top-3 teams, and we deem this variation as acceptable.
  
The top-3 teams in the final round of the DC track and the top-3 in the final phase of the DD track were awarded with bonus. Apart from this, the top-3 teams in each of the two precedent rounds of the DC track were also awarded with some bonus, to encourage early submissions of face-swap data. 

\section{Datasets}
As shown in Fig. \ref{fig_design}, there are four datasets in our competition, which are the resource dataset, the public dataset, the private dataset 1, and the private dataset 2. We describe each of these datasets in more details in the following.

\subsection{The Creation Dataset}
The resource dataset was used in the DC track to create face-swap clips based on paired resource clips. To construct this dataset, we first collected 40 recorded videos of 40 consented subjects. Considering that high-quality DeepFakes should be able to handle varies facial expressions, in the recording, we asked each subject to imitate the facial performance and spoken lines of some classic movie footage. Each movie footage is about one minute long, and the subject can see the laptop screen playing the footage and imitate the performance line by line, during which the subject can pause the movie freely. We then cut out the intermediate pauses of the subject video and saved the valid imitation performances to shorter clips to form each subject's resource clips. 

\begin{figure}[ht]
\centering
\centerline{\includegraphics[width=0.4\textwidth]{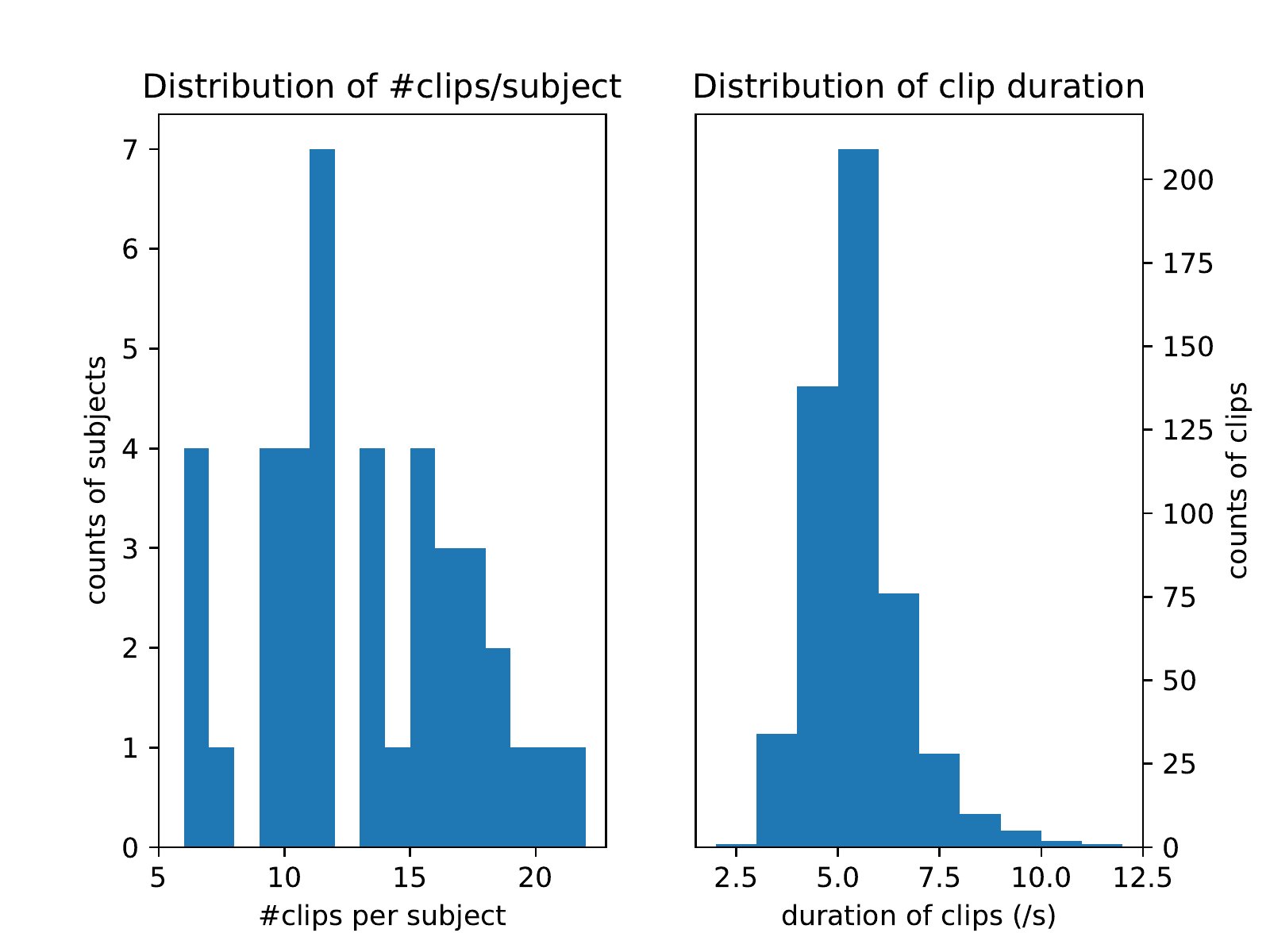}}
\caption{Statistics of the resource dataset.}
\label{fig_resource}
\end{figure}

In general, each resource clip contains one complete line spoken by the subject. In total, we have 505 clips in the resource dataset. The distribution of clip duration and the distribution of the number of  clips for each subject can be seen in Fig. \ref{fig_resource}. The 40 subjects are organized into 20 groups, and the two subjects in each group are to be face-swapped in a mutual way. To avoid odd and unnatural swap results, we matched the gender, ethnic group, and age between the two subjects in each group. The recruited subjects are 50-50 in gender, and the proportion of ethnic groups is 50\% Asian, 25\% Caucasian and 25\% African. They were informed with the usage, paid with rewards, and signed consents for data collection and research usage. The two subjects in each group imitated the same movie footage to cover roughly the same range of expressions and spoken lines to facilitate creating better face swap effects. The resource videos were captured in various lighting conditions, but the two subjects in the same group were captured in the same condition to facilitate better face swap effects.
%

\subsection{The Detection Datasets}
The public test set and the private-1 test set were constructed from participants-submitted face-swap clips and real clips cut from the resource videos. There were three submission rounds in the DC track. In each round, a different set of 80 face-swap clips were to be submitted in a submission, composing 2 clips for each of the 40 subject. The number of DC track submissions (including the organizers' baseline submission), the methods/sources of the face-swap clips, and other details of the datasets are shown in Table \ref{tab_Datasets}. As can be seen, there were totally 35 DC track submissions that contributed to the face-swap clips in the public set and the private-1 set. There were 7 different methods used to create these face-swap clips, from the participants-reported information on their solutions. They also used various post-processing operations to improve the video quality or to obtain better anti-detection performance. Besides, we also conducted random post-processing operations selected from FFmpeg compression quality c40 (heavy compression),  c23 (light compression), and original (no compression) to both fake and real clips, to imitate real-world video quality degradation on social network \cite{FaceForensics++}.
Note that the public set and the private-1 set both contain clips from 20 subjects, but they are disjoint IDs. 

\begin{figure*}[ht]
\centering
\centerline{\includegraphics[width=0.85\textwidth]{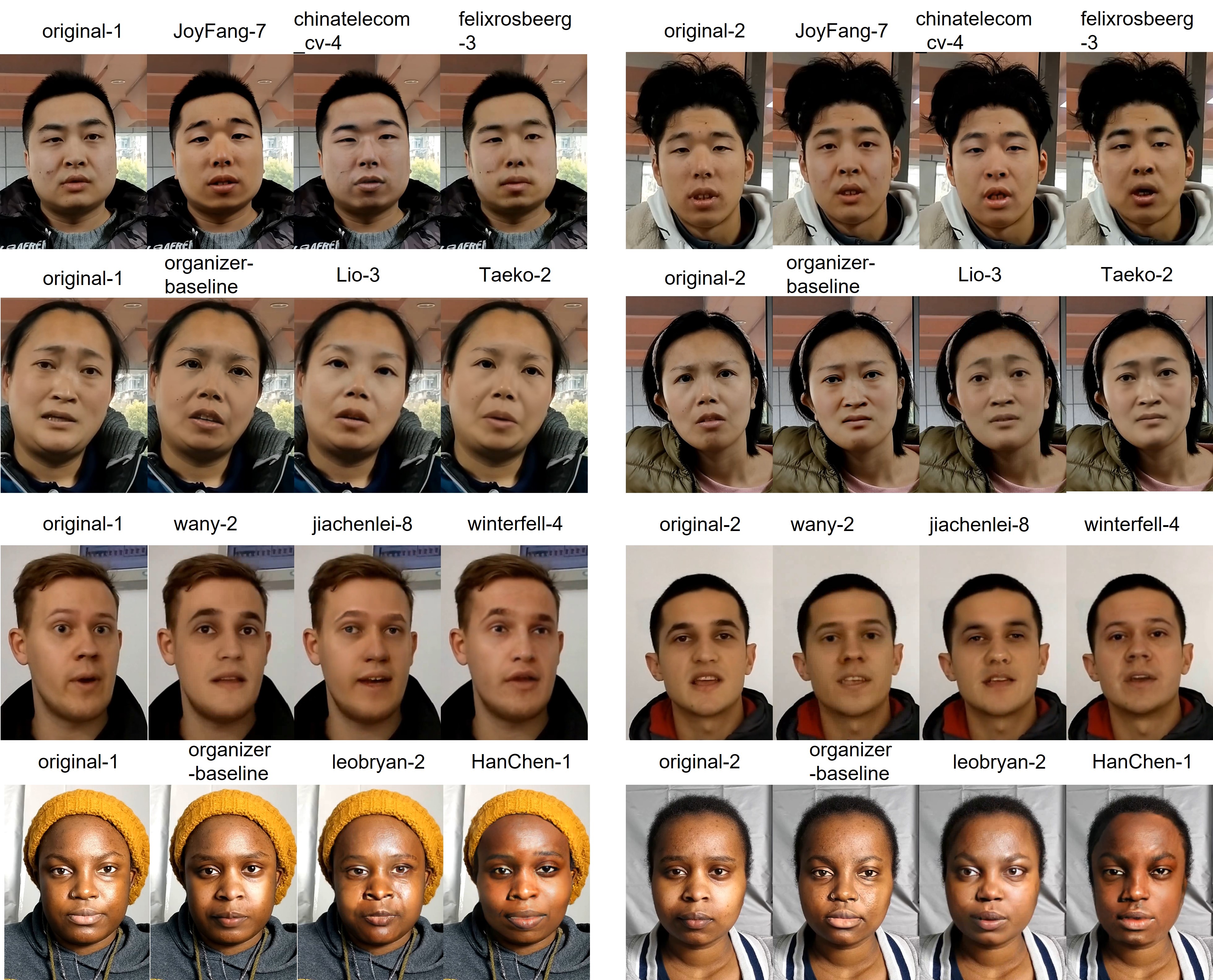}}
\caption{Sample face crops of the DFGC-2022 dataset. The \textit{original-1} and \textit{original-2} faces in each row are a pair to be face-swapped mutually, the caption above each face-swap image is the identifier of the submission in the creation track that included this face-swap clip.}
\label{fig_dataset}
\end{figure*}

The numbers of real and fake video clips in these datasets can also be seen in Table \ref{tab_Datasets}. In total, we built a face-swap DeepFake dataset containing 4394 video clips, together with our participants. This dataset features diverse face-swap methods and post-processing operations, and we release this dataset for future DeepFake related research. Sample images of this dataset can be seen in Fig. \ref{fig_dataset}.
\begin{table} [th] 
\centering
\caption{Details of the competition datasets.}  \label{tab_Datasets}
\renewcommand{\arraystretch}{1.0}
\scalebox{0.75}{%
\begin{tabular}{|c|c|c|c|}
  \hline
        & Public Set   & Private-1 Set     & Private-2 Set      \\
  \hline
\makecell{\# DC track \\submissions}  &35  &35  &/   \\
\hline
\# IDs  &20   &20   &/ \\
  \hline
\makecell{Methods/\\Sources}  &\makecell{DeepFaceLab \cite{DeepFaceLab},\\ SimSwap \cite{SimSwap}, \\FaceShifter \cite{FaceShifter}, \\FaceSwapper \cite{FaceSwapper}, \\MegaFS \cite{MegaFS}, \\ InfoSwap \cite{InfoSwap}, \\Self-proposed}  &\makecell{Same as \\Public Set} &\makecell{ZAO, \\FaceMagic, \\ReFace, \\Jiggy, \\YouTube-DF \cite{YouTube-DF}}  \\
\hline
\makecell{DC Participants’ \\post-processings}  &\makecell{Compression,\\ Super-resolution, \\De-blurring, \\Adversarial Noise}
  &\makecell{Same as \\Public Set}  &/ \\
  \hline
\makecell{Organizers’ \\post-processings}     &\makecell{Original, \\c23, c40}     &\makecell{Same as \\Public Set}     &/ \\
\hline
\# Real clips    &829    &766    &424 \\
\hline
\# Fake clips    &1399   &1400   &471 \\
\hline
\# All clips     &2228   &2166   &895 \\
\hline
\end{tabular} 
}
\end{table}

To include more real-world face-swap and real data in the final phase of the DD track, we devised the private-2 test set. Details of this dataset can also be seen in Table \ref{tab_Datasets}. We collected and crawled some face-swap clips and the corresponding real clips from some popular apps, i.e. ZAO, FaceMagic, ReFace, and Jiggy. We also included the data from the YouTube-DF \cite{YouTube-DF} dataset, which collected real-world face-swap videos from YouTube. The real and fake video clips in the private-2 set mostly depict film and television scenes, and hence it complements the private-1 set which depicts relatively controlled indoor scenes. The private-1 set and private-2 set were combined together to form the test dataset for the final phase of the DD track.
\section{Evaluation Methods}
We designed comprehensive evaluation methods for both the creation track and the detection track. The evaluation of DeepFake creation submissions is multifaceted, because a good DeepFake video should meet multiple requirements at the same time. Considering that DeepFake attacks often target at deceiving both human perception and machine perception (detection), we devised both subjective and objective evaluation metrics respectively for the two forms of perception. Our subjective metrics include evaluations of the face-swap clips in 5 aspects, i.e., overall realism, video quality, facial expression, mouth movement, and facial ID. The objective metric evaluates the anti-detection ability against DeepFake detection models. 

We then elaborate on the evaluation metrics and methods for the DeepFake creation submissions. We first recruited 5 human raters to form the subjective evaluation board, who are all graduate students from our lab who are very familiar with DeepFakes. They were told to independently rate each face-swap clip from the aforementioned 5 aspects on a discrete scale of 1 (very bad) to 5 (very good). Their ratings are then averaged to obtain the subjective scores. The overall realism score ($s_r$) represents the perceptual realism after viewing the complete video clip, where both spatial and temporal artifacts may be noticed by the raters. The video quality score ($s_q$) represents the  closeness of the face-swap result video quality compared to the original target video quality. Post-processing operations like video compression may influence this score. The facial expression score ($s_e$) represents the closeness of result facial expression compared to the original target video. The mouth movement score ($s_m$) represents the consistency of the result mouth movement with the spoken audio. The facial ID score ($s_{id}$) represents the closeness of the result facial ID to the source ID. Here, the target and the source are respectively the background video swapped onto and the donor ID swapped from in the face-swap vocabulary. The realism $s_r$ and the mouth movement $s_m$ are directly scored based only on the result video, while the other three aspects need to be scored by comparing with either the target or the source video as the reference. During the subjective scoring, the same result videos submitted by all different participants are viewed one after another, for a better sense of comparison between different submissions. 

To calculate the objective anti-detection score, the result video will be detected by the top-3 detection track models to calculate the average detection score ($s_d$) in the range of [0, 1], where 1 is the largest possibility of predicting as fake. The anti-detection sore is then $1-s_d$. For the first two rounds in the creation track, since the final top-3 detection models were not determined yet, the anti-detection scores of the submissions in those rounds were obtained using top detection models at that time.
Finally, the overall score for a face-swap video clip is calculated as follows:
\begin{equation}
s = s_r + s_q + s_e + s_m + s_{id} + 10(1-s_d)
\end{equation}
After balancing, the five subjective scores each have a range of 5 points while the anti-detection score has a range of 10 points. The final score for a DeepFake creation submission is the average score of all evaluated video clips in it.

As for the evaluation of the detection track submissions, we use the Area Under Curve (AUC) score of the ROC curve. The AUC metric is robust to unbalanced real and fake data as in our case.
\section{Results}
We then report the final results of both the creation track and the detection track. For the DeepFake creation track, there were a total of 35 submissions from 11 teams during the three submission rounds. 
The full results of all submissions can be seen in Table \ref{tab_C_solutions}. Note that submission scores from different rounds are not directly comparable, since the used resource clips and detection models are all different.
According to the sum score, the teams \textit{JoyFang, 
chinatelecom\_cv,} and \textit{felixrosberg} achieved Top-3 in the final round of the DeepFake creation track. 

In the DeepFake detection track, a total of 25 teams made submissions to the validation-phase leaderboard. Top-10 teams on this leaderboard (which was later extended to Top-13) were requested to send their inference codes and dockers to the organizers to complete the final phase test on the private test set, and 8 teams fed back at last. The AUC-ROC results, as well as False Positive Rate (FPR), False Negative Rate (FNR), and 0.5-prediction rate (0.5-rate), of all final phase models are listed in Table \ref{tab_D_solutions}. Here, positive predictions are those greater than 0.5, negative ones less than 0.5, and most 0.5-predictions result from unhandled or erroneous conditions, e.g. no faces detected. The results on the public set, the private-1 set and the private-2 set are shown, and the final ranking is determined by the AUC-ROC result on the whole private test set. The teams \textit{HanChen, OPDAI,} and \textit{guanjz} achieved Top-3 in the DeepFake detection track.

The top-3 detection teams are all from different affiliations from those of the top-3 creation teams. Participants from the organizers' affiliations are allowed as long as they are not directed by the organizers.

In Table \ref{tab_C_solutions} and \ref{tab_D_solutions}, we also summarize and compare the submissions from a high-level technical perspective. These summarizations are based on self-reported information collected from our participants. Note that the technical summarizations in these tables are rather sketchy, because of different abstraction levels of the original collected information and the limited space. Technical details like hyper-parameters, tricks and their different combinations may make big differences. Hence, we should interpret them with caution. Still, some relatively common observations can be obtained. For example, improved merging and careful post-processing techniques are important to achieve better performance in face-swap creation. In the detection track, more and diverse training datasets seem to be more effective. Most detection models have higher FNR than FPR, implying that unseen types of fake data are often missed and pose a serious challenge for current methods.
\begin{table*}[th]
\caption{Summary comparison of all creation track submissions. Columns 2-8 show the scores of each submission. Columns 10-11 summarize the used face-swap models and post-processing in creation.}  \label{tab_C_solutions}
\renewcommand{\arraystretch}{1.0}
\scalebox{0.7}{%
\begin{tabular}{lllllllllll}
\hline
submission\_ID       & realism & mouth & quality & expression & ID    & Anti-Det & sum    & Round & model$^a$ & post-process$^b$ \\ \hline
organizer-baseline-1 & 3.545   & 3.99  & 3.88    & 3.7        & 3.775 & 2.525    & 21.415 & 1     & DFL    &              \\
JoyFang-2            & 3.24    & 4.02  & 3.69    & 3.55       & 3.055 & 3.172    & 20.727 & 1     & DFL    & IM           \\
JoyFang-4            & 2.96    & 3.99  & 3.005   & 3.635      & 3.055 & 2.043    & 18.688 & 1     & DFL    & IM+AN        \\
chinatelecom\_cv-1   & 2.73    & 4.025 & 3.105   & 3.58       & 2.93  & 2.498    & 18.868 & 1     & SS     &              \\
jiachenlei-1         & 2.675   & 3.91  & 2.945   & 3.4        & 2.78  & 2.367    & 18.077 & 1     & SS     &              \\
leobryan-1           & 1.93    & 3.67  & 2.565   & 3.13       & 2.465 & 2.665    & 16.425 & 1     & FSh    &              \\ \hline
chinatelecom\_cv-3   & 3.895   & 4.055 & 4.115   & 3.875      & 3.245 & 7.774    & 26.959 & 2     & SS     & SR+IM        \\
JoyFang-5            & 3.72    & 3.995 & 3.95    & 3.825      & 3.72  & 6.679    & 25.889 & 2     & DFL    & IM           \\
organizer-baseline-2 & 3.845   & 4.05  & 4.025   & 3.89       & 3.88  & 5.787    & 25.477 & 2     & DFL    &              \\
JoyFang-6            & 3.91    & 4.01  & 3.95    & 3.855      & 3.655 & 6.034    & 25.414 & 2     & DFL    & IM+AN        \\
chinatelecom\_cv-2   & 3.09    & 3.755 & 3.695   & 3.545      & 3.155 & 6.761    & 24.001 & 2     & SS     &              \\
jiachenlei-2         & 3.215   & 3.72  & 3.645   & 3.625      & 2.93  & 6.683    & 23.818 & 2     & SS     & IM           \\
Lio-1                & 3.65    & 3.965 & 3.665   & 3.84       & 3.495 & 5.12     & 23.735 & 2     & DFL    &              \\
jiachenlei-3         & 3.445   & 3.78  & 2.73    & 3.705      & 2.89  & 6.823    & 23.373 & 2     & SS     & IM+Co         \\
winterfell-3         & 2.93    & 3.645 & 3.645   & 3.325      & 2.785 & 6.421    & 22.751 & 2     & FSw    &              \\
leobryan-2           & 2.63    & 3.425 & 3.39    & 3.195      & 2.755 & 6.774    & 22.169 & 2     & SS     &              \\
HanChen-1            & 1.795   & 2.91  & 3.28    & 2.83       & 2.61  & 7.539    & 20.964 & 2     & IS     &              \\
winterfell-1         & 2.755   & 3.62  & 3.15    & 3.49       & 2.925 & 4.948    & 20.888 & 2     & SS     &              \\
HanChen-2            & 1.775   & 2.91  & 3.285   & 2.82       & 2.55  & 6.535    & 19.875 & 2     & IS     & IM           \\ \hline
JoyFang-7            & 4.04    & 4.095 & 4.16    & 3.89       & 3.6   & 5.04     & 24.825 & 3     & DFL    & IM+DB        \\
chinatelecom\_cv-4   & 3.79    & 4.05  & 4.15    & 3.91       & 3.065 & 5.106    & 24.071 & 3     & SS+IT  & IM           \\
JoyFang-8            & 3.97    & 4.08  & 3.995   & 3.905      & 3.61  & 4.441    & 24.001 & 3     & DFL    & IM+DB+AN     \\
chinatelecom\_cv-5   & 3.555   & 3.925 & 4.075   & 3.815      & 2.96  & 4.895    & 23.225 & 3     & SS+IT  & SR+IM        \\
felixrosberg-3       & 3.555   & 3.94  & 3.825   & 3.78       & 3.15  & 4.093    & 22.343 & 3     & NM     & SR           \\
felixrosberg-2       & 3.48    & 3.965 & 3.85    & 3.715      & 3.185 & 4.092    & 22.287 & 3     & NM     &              \\
organizer-baseline-3 & 3.79    & 4.015 & 4.03    & 3.88       & 3.84  & 2.625    & 22.18  & 3     & DFL    &              \\
Lio-3                & 3.915   & 4.07  & 3.79    & 3.88       & 3.715 & 2.762    & 22.132 & 3     & DFL    & IM+AN        \\
Taeko-2              & 3.72    & 3.945 & 3.98    & 3.735      & 3.435 & 3.313    & 22.128 & 3     & DFL+IT & IM           \\
wany-2               & 3.955   & 4.055 & 3.7     & 3.88       & 3.625 & 2.695    & 21.91  & 3     & DFL    &              \\
Lio-2                & 3.87    & 4.09  & 3.905   & 3.865      & 3.735 & 2.386    & 21.851 & 3     & DFL    & IM           \\
wany-1               & 3.96    & 4.085 & 3.905   & 3.885      & 3.62  & 2.329    & 21.784 & 3     & DFL    & IM           \\
jiachenlei-8         & 2.6     & 3.32  & 3.245   & 3.085      & 2.625 & 2.633    & 17.508 & 3     & MFS & Co            \\
winterfell-4         & 3.17    & 3.625 & 3.24    & 3.5        & 2.945 & 0.767    & 17.247 & 3     & SS     &              \\
jiachenlei-7         & 2.51    & 3.295 & 3.54    & 3.03       & 2.61  & 2.095    & 17.08  & 3     & MFS &              \\
winterfell-5         & 1.805   & 2.88  & 3.385   & 2.67       & 2.195 & 0.695    & 13.63  & 3     & MFS &              \\ \hline
\end{tabular}
}

\scriptsize{$^a$ DFL: DeepFaceLab, SS: SimSwap, FSh: FaceShifter, FSw: FaceSwapper, IS: InfoSwap, NM: New Model, MFS: MegaFS, IT: Improved Training.
} \\
\scriptsize{$^b$ IM: Improved Merging, AN: Adversarial Noise, SR: Super Resolution, Co: Compression, DB: De-blurring. 
}
\end{table*}
\section{Top Solutions}
We introduce the overall ideas of top-3 solutions in both the creation track and the detection track in more detail.  The following are descriptions of the solutions of the top-3 creation teams.
\begin{itemize}
    \item The \textit{JoyFang} team's best solution (\textit{JoyFang-7}) employed the DeepFaceLab \cite{DeepFaceLab} tool with improved adaptive merging that is better at light adapting. They then used deblurring as post-processing. 
    \item The \textit{chinatelecom\_cv} team's best solution \textit{chinatelecom\_cv-4} improved the SimSwap \cite{SimSwap} method by adding local discriminators on the eyes and mouth regions. Based on the original loss, they also adopted local discrimination loss, VGGFace loss, and total variation loss. The model was trained on the CelebA-HQ dataset. In the post-processing, they only fused the eyes, nose and mouth regions and used the Gaussian filter on the margins. They also used a self-developed face super-resolution model to improve subjective quality.
    \item The \textit{felixrosberg} team's best solution \textit{felixrosberg-3} was a new face-swap model proposed by them, which is not published yet, in $256 \times 256$ resolution which was then post-processed with super-resolution to $512 \times 512$.
\end{itemize}
Then, the following are descriptions of the top-3 detection team's solutions.
\begin{itemize}
    \item The \textit{HanChen} team used an ensemble of three models, two ConvNext \cite{convNext} (convnext\_xlarge\_384\_in22ft1k) at different epoches and one Swin-Transformer \cite{swin} (swin\_large\_patch4\_window12\_384\_in22k), both initialized with weights pretained on the ImageNet dataset. Their training datasets include FaceForensics++ (Deepfakes, Face2Face, FaceSwap, FaceShifter, NeuralTextures, DeepFakeDetection (DFD)) \cite{FaceForensics++} in c23 and c40 qualities, HifiFace \cite{Hififace} on FaceForensics++, UADFV \cite{UADFV}, DF-TIMIT \cite{DF-TIMIT}, DeeperForensics-1.0 \cite{DeeperForensics}, Celeb-DF \cite{CelebDF}, WildDeepfake (WDF) \cite{WildDF}, and DFDC \cite{DFDC}. The models were trained on cropped $384 \times 384$ face images with BCE loss. Data augmentations include HorizontalFlip, GaussNoise, and GaussianBlur.
    \item The \textit{OPDAI} team used an ensemble of five models, including Swin-Transformer \cite{swin}, EfficientNet-B7 (EffNet-B7) \cite{efficientnet}, and EfficientNet-B0. Their training datasets include DFDC \cite{DFDC}, Celeb-DF \cite{CelebDF}, and FaceForensics++ \cite{FaceForensics++} (Deepfakes, Face2Face, FaceSwap, and NeuralTextures). Their models were trained on cropped face images, and one model fuses predictions from key facial parts like nose, eyes and mouth. They used data augmentations including Compression, GaussNoise, GaussianBlur, HorizontalFlip, Adjustment Brightness/Contrast/HueSaturationValue, Rotate, and facial parts erasing. For different models, they used different training losses and strategies, which include BCE loss, balanced MSE loss, BCE loss on feature maps, label smoothing, and online hard example mining.
    \item The \textit{guanjz} team used an ensemble of five models, including two Vision Transformers \cite{ViT} (w/ and w/o dropout layer), two Swin-Transformers \cite{swin} (w/ and w/o dropout layer), and one temporal-aware model TSM \cite{TSM} based on Resnet50 features. Their training datasets includes FaceForensics++ (with FaceShifter and DeepFakeDetection) \cite{FaceForensics++}, DFDC \cite{DFDC}, Celeb-DF \cite{CelebDF}, DeeperForensics \cite{DeeperForensics}, Kodf \cite{Kodf}, and ForgeryNet \cite{ForgeryNet}. Data augmentations include compression, down scale, horizontal flip, random shift, random scale, random rotate, Gaussian blur, and random color change.
\end{itemize}
\begin{table*}[th]
\caption{Summary comparison of all final-phase detection track submissions. Columns 2-5 show AUC-ROC/FPR/FNR/0.5-rate on different testing sets. Columns 6-11 summarize the detection model architectures, the training datasets, and data augmentation of each solution.}  \label{tab_D_solutions}
\renewcommand{\arraystretch}{1.0}
\scalebox{0.7}{%
\begin{tabular}{|l|l|l|l|l|l|l|l|l|l|l|l|}
\hline
Team          & Public & Private-1 & Private-2 & Private & Model Architectures                                      & \makecell{\#Models \\Ensembled} & Datasets                                                         & \#Datasets & Augmentation                   & \#Aug$^c$ \\
\hline
HanChen       & \makecell{0.9521 \\5.5\% \\17.4\% \\0.0\%} & \makecell{0.9178 \\10.8\% \\ 21.4\% \\ 0.0\%}    &\makecell{0.8955 \\16.0\% \\25.1\% \\0.0\%}    & \makecell{0.9085 \\12.7\% \\22.3\% \\0.0\%}  & \makecell{convnext\_xlarge\_384, \\swin\_large\_384}       & 3        &\makecell{ FF++, HifiFace, \\UADFV, DF-TIMIT, \\DeeperForensics, \\DFD, CelebDF, \\WDF, DFDC} & 9          & Fl, GN, GB                       & 3                            \\ 
\hline
OPDAI         & \makecell{0.9297 \\2.2\% \\39.0\% \\0.0\%} & \makecell{0.8836 \\6.8\% \\35.4\% \\0.0\%}    & \makecell{0.8511 \\8.5\% \\20.6\% \\23.0\%}    & \makecell{0.8672 \\7.4\% \\31.7\% \\6.7\%}  & \makecell{swin\_large\_384, \\EffNetB7, EffNetB0}           & 5        & \makecell{DFDC, CelebDF, FF++}                                                 & 3          & \makecell{Co, GN, GB, \\Fl, Cl, Ro, \\Er}           & 7                         \\ 
\hline
guanjz        & \makecell{0.9483 \\12.2\% \\13.8\% \\0.0\%} & \makecell{0.911 \\18.9\% \\17.0\% \\0.0\%}     & \makecell{0.7461 \\42.2\% \\26.1\% \\1.1\%}    & \makecell{0.867 \\ 27.2\% \\19.3\% \\0.3\%}   & \makecell{vit\_small\_384, \\swin\_base\_384, \\resnet50-TSM} & 5        &\makecell{ FF++, DFD, DFDC, \\CelebDF, \\DeeperForensics, \\Kodf, ForgeryNet}             & 7          & \makecell{Co, Sc, Fl, \\Sh, Ro, GB, \\Cl}           & 7                       \\ 
\hline
jiachenlei    & \makecell{0.8666 \\2.9\% \\48.9\% \\0.0\%} & \makecell{0.8746 \\0.8\% \\50.6\% \\0.0\%}    & \makecell{0.874 \\5.0\% \\44.8\% \\0.0\%}     & \makecell{0.8665 \\2.3\% \\49.1\% \\0.0\%}  & \makecell{EffNetB4}                                     & 2        & \makecell{FF++, DFDC}                                                               & 2         & \makecell{Co, GN, GB, \\Fl, Cl, DO, \\Sh, Ro, Sc
}                              & 9                         \\
\hline
wany          & \makecell{0.9078 \\4.6\% \\42.8\% \\0.3\%} & \makecell{0.8552 \\14.8\% \\29.9\% \\0.4\%}    & \makecell{0.8408 \\26.9\% \\22.3\% \\0.0\%}    & \makecell{0.8474 \\19.1\% \\28.0\% \\0.3\%}  & \makecell{vit\_base\_384, \\swin\_large\_384}              & 2        & \makecell{CelebDF, DFD, \\FSh, MegaFS}                                          & 4          & \makecell{GN, GB, Cl, \\Shp, ISON}              & 5                            \\ 
\hline
leobryan      & \makecell{0.9565 \\0.4\% \\62.8\% \\0.0\%} & \makecell{0.8644 \\2.5\% \\71.3\% \\0.0\%}    & \makecell{0.8132 \\7.1\% \\69.0\% \\0.0\%}    & \makecell{0.845 \\4.1\% \\70.7\% \\0.0\%}   & \makecell{EffNetB0, EffNetB3}                            & 6        & \makecell{FF++, DFDC-preview}                                                        & 2          & \makecell{Co, Fl, GN, \\GB, DO, Shp, \\Sh, Sc, Ro, \\Er} & 10                      \\ 
\hline
xwj           & \makecell{0.9618 \\0.1\% \\ 46.9\% \\0.0\%} & \makecell{0.9178 \\1.3\% \\69.1\% \\0.0\%}    & \makecell{0.7207 \\16.7\% \\62.4\% \\0.0\%}    & \makecell{0.8352 \\6.8\% \\67.4\% \\0.0\%}  & \makecell{EffNetB7-ViT}\cite{Eff-vit}                                 & 7        & \makecell{DFDC, CelebDF, FF++}                                                 & 3          &\makecell{ Co, GN, GB, \\Sc, Cl}                 & 5                            \\ 
\hline
walk\_at\_zoo & \makecell{0.9599 \\1.8\% \\22.9\% \\0.0\%} & \makecell{0.9314 \\9.3\% \\19.4\% \\0.0\%}    & \makecell{0.4507 \\83.0\% \\8.3\% \\4.4\%}    & \makecell{0.7922 \\35.5\% \\16.6\% \\1.3\%}  & \makecell{EffNetB0}                                     & 5        & \makecell{FF++, CelebDF, \\DFDC-preview, DFDC, \\ForgeryNet}                                & 5          & \makecell{Co, Fl, Cl, \\DO, Shp, GN, \\GB, Er}       & 8                       \\ \hline
\end{tabular}
}
\\
\scriptsize{$^c$ Co: compression, Sc: scale, Fl: flip, Sh: shift, Ro: rotation, GB: Gaussian Blur, Cl:	Color changes (including intensity, hue, contrast etc.), GN: Gaussian Noise, Er: Part Erasing, Shp: sharpen, ISON: ISONoise, DO: DropOut.
} 
\end{table*}
\section{Conclusions}
In this paper, we report the design,  datasets, evaluation, results and top solutions of the DFGC-2022 competition. It organized the deepfake creation and detection sides in the same competition to combat with each other. Developing on its precedent DFGC-2021 competition, it built a more advanced video deepfake dataset, more subjective metrics were considered for evaluating human perception of created face-swap clips, in-the-wild data sources were additionally introduced in the final evaluation of detection models. All these improvements make it a more realistic game compared with its precedent. 

From the competition results and participants-reported information about their solutions, we summarize the following  key observations:
\begin{itemize}
\item Careful post-processing operations were adopted by the creation track participants to either improve the deception ability to humans (e.g. de-blurring, super-resolution, advanced blending or merging methods) or to detection models (e.g. compression). Different from in the DFGC-2021, only four submissions added adversarial noise in this competition, and their effects on anti-detection scores seem minor. This may be because video coding and compression have negative impacts on the effects of adversarial noise.
\item 
In the final creation round, the highest scores for realism, mouth, quality, expression, and ID are respectively 4.04, 4.095, 4.16, 3.91, 3.84 for a full score of 5. This implies that the overall realism, the matchiness of mouth and words, and the video quality of best face-swap solutions can already achieve good performances, while there is still room for improvements for the accuracy of expression and ID similarity as perceived by humans. As for deceiving detection models, there were two face-swap solutions that achieved scores higher than $5.0$, meaning that they can be detected as real by the top-3 detectors on average.
\item Calculating the correlation coefficients between different metrics using all submission scores in the final creation round, we observe high mutual correlations between realism, mouth, and expression scores (higher than 0.99), the correlation of them with the quality score is lower (around 0.78), and that with the anti-detection score is further lower (around 0.60).
\item Top detection solutions all used ensembles of most recent effective deep models, e.g. vision transformers. They also trained the models on many deepfake datasets and used various data augmentations.
\item Top detection models still suffer from generalization problems. The high performances degraded when changing the test dataset from the public set to the private-1 and especially the private-2 dataset, implying that there is still a large space of improvement for the detection generalization.
\end{itemize}

The adversarial game between deepfake creation and its detection is an evolutionary course that will last for a long time. For the current state, best detection models still struggle when facing the best quality deepfake videos and suffer from generalization problems. We hope to carry on with the DFGC competition series in the future to closely monitor the progresses in both sides.
\section*{Acknowledgements}
We thank the Tianjin Academy for Intelligent Recognition Technology for sponsoring the competition awards. We also thank our participating teams for agreeing to report descriptions of their solutions.
{\small
\bibliographystyle{ieee}
\bibliography{egbib}

\begin{thebibliography}{10}\itemsep=-1pt

\bibitem{url_faceswap}
{deepfakes/faceswap}.
\newblock \url{https://github.com/deepfakes/faceswap}.

\bibitem{url_DFDC1}
{DFDC 1st Place Solution}.
\newblock \url{https://github.com/selimsef/dfdc_deepfake_challenge}.

\bibitem{url_DFDC2}
{DFDC 2nd Place Solution}.
\newblock \url{https://github.com/cuihaoleo/kaggle-dfdc}.

\bibitem{url_DeepFaceLab}
{iperov/DeepFaceLab}.
\newblock \url{https://github.com/iperov/DeepFaceLab}.

\bibitem{url_deepfake}
{wiki-Deepfake}.
\newblock \url{https://en.wikipedia.org/wiki/Deepfake}.

\bibitem{lip_match}
S.~Agarwal, H.~Farid, O.~Fried, and M.~Agrawala.
\newblock Detecting deep-fake videos from phoneme-viseme mismatches.
\newblock In {\em Proceedings of the IEEE/CVF conference on computer vision and
  pattern recognition workshops}, pages 660--661, 2020.

\bibitem{world_leader}
S.~Agarwal, H.~Farid, Y.~Gu, M.~He, K.~Nagano, and H.~Li.
\newblock Protecting world leaders against deep fakes.
\newblock In {\em CVPR workshops}, volume~1, page~38, 2019.

\bibitem{SimSwap}
R.~Chen, X.~Chen, B.~Ni, and Y.~Ge.
\newblock Simswap: An efficient framework for high fidelity face swapping.
\newblock In {\em Proceedings of the 28th ACM International Conference on
  Multimedia}, pages 2003--2011, 2020.

\bibitem{Eff-vit}
D.~A. Coccomini, N.~Messina, C.~Gennaro, and F.~Falchi.
\newblock Combining efficientnet and vision transformers for video deepfake
  detection.
\newblock In {\em International Conference on Image Analysis and Processing},
  pages 219--229. Springer, 2022.

\bibitem{On_the_detection}
H.~Dang, F.~Liu, J.~Stehouwer, X.~Liu, and A.~K. Jain.
\newblock On the detection of digital face manipulation.
\newblock In {\em Proceedings of the IEEE/CVF Conference on Computer Vision and
  Pattern recognition}, pages 5781--5790, 2020.

\bibitem{DFDC}
B.~Dolhansky, J.~Bitton, B.~Pflaum, J.~Lu, R.~Howes, M.~Wang, and
  C.~Canton-Ferrer.
\newblock The deepfake detection challenge dataset.
\newblock {\em ArXiv}, abs/2006.07397, 2020.

\bibitem{ViT}
A.~Dosovitskiy, L.~Beyer, A.~Kolesnikov, D.~Weissenborn, X.~Zhai,
  T.~Unterthiner, M.~Dehghani, M.~Minderer, G.~Heigold, S.~Gelly, et~al.
\newblock An image is worth 16x16 words: Transformers for image recognition at
  scale.
\newblock {\em arXiv preprint arXiv:2010.11929}, 2020.

\bibitem{InfoSwap}
G.~Gao, H.~Huang, C.~Fu, Z.~Li, and R.~He.
\newblock Information bottleneck disentanglement for identity swapping.
\newblock In {\em Proceedings of the IEEE/CVF Conference on Computer Vision and
  Pattern Recognition}, pages 3404--3413, 2021.

\bibitem{ForgeryNet}
Y.~He, L.~Sheng, J.~Shao, Z.~Liu, Z.~Zou, Z.~Guo, S.~Jiang, C.~Sun, G.~Zhang,
  K.~Wang, et~al.
\newblock Forgerynet--face forgery analysis challenge 2021: Methods and
  results.
\newblock {\em arXiv preprint arXiv:2112.08325}, 2021.

\bibitem{see_better}
T.~Hu, H.~Qi, Q.~Huang, and Y.~Lu.
\newblock See better before looking closer: Weakly supervised data augmentation
  network for fine-grained visual classification.
\newblock {\em arXiv preprint arXiv:1901.09891}, 2019.

\bibitem{DeeperForensics}
L.~Jiang, R.~Li, W.~Wu, C.~Qian, and C.~C. Loy.
\newblock Deeperforensics-1.0: A large-scale dataset for real-world face
  forgery detection.
\newblock In {\em Proceedings of the IEEE/CVF conference on computer vision and
  pattern recognition}, pages 2889--2898, 2020.

\bibitem{StyleGAN}
T.~Karras, S.~Laine, and T.~Aila.
\newblock A style-based generator architecture for generative adversarial
  networks.
\newblock In {\em Proceedings of the IEEE/CVF conference on computer vision and
  pattern recognition}, pages 4401--4410, 2019.

\bibitem{DF-TIMIT}
P.~Korshunov and S.~Marcel.
\newblock Vulnerability assessment and detection of deepfake videos.
\newblock In {\em 2019 International Conference on Biometrics (ICB)}, pages
  1--6. IEEE, 2019.

\bibitem{YouTube-DF}
I.~Kukanov, J.~Karttunen, H.~Sillanp{\"a}{\"a}, and V.~Hautam{\"a}ki.
\newblock Cost sensitive optimization of deepfake detector.
\newblock In {\em 2020 Asia-Pacific Signal and Information Processing
  Association Annual Summit and Conference (APSIPA ASC)}, pages 1300--1303.
  IEEE, 2020.

\bibitem{Kodf}
P.~Kwon, J.~You, G.~Nam, S.~Park, and G.~Chae.
\newblock Kodf: A large-scale korean deepfake detection dataset.
\newblock In {\em Proceedings of the IEEE/CVF International Conference on
  Computer Vision}, pages 10744--10753, 2021.

\bibitem{center_loss}
J.~Li, H.~Xie, J.~Li, Z.~Wang, and Y.~Zhang.
\newblock Frequency-aware discriminative feature learning supervised by
  single-center loss for face forgery detection.
\newblock In {\em Proceedings of the IEEE/CVF conference on computer vision and
  pattern recognition}, pages 6458--6467, 2021.

\bibitem{FaceShifter}
L.~Li, J.~Bao, H.~Yang, D.~Chen, and F.~Wen.
\newblock Faceshifter: Towards high fidelity and occlusion aware face swapping.
\newblock {\em arXiv preprint arXiv:1912.13457}, 2019.

\bibitem{face-Xray}
L.~Li, J.~Bao, T.~Zhang, H.~Yang, D.~Chen, F.~Wen, and B.~Guo.
\newblock Face x-ray for more general face forgery detection.
\newblock In {\em Proceedings of the IEEE/CVF conference on computer vision and
  pattern recognition}, pages 5001--5010, 2020.

\bibitem{FaceSwapper}
Q.~Li, W.~Wang, C.~Xu, and Z.~Sun.
\newblock Learning disentangled representation for one-shot progressive face
  swapping.
\newblock {\em arXiv preprint arXiv:2203.12985}, 2022.

\bibitem{blinking}
Y.~Li, M.-C. Chang, and S.~Lyu.
\newblock In ictu oculi: Exposing ai created fake videos by detecting eye
  blinking.
\newblock In {\em 2018 IEEE International workshop on information forensics and
  security (WIFS)}, pages 1--7. IEEE, 2018.

\bibitem{UADFV}
Y.~Li, M.-C. Chang, and S.~Lyu.
\newblock In ictu oculi: Exposing ai created fake videos by detecting eye
  blinking.
\newblock In {\em 2018 IEEE International Workshop on Information Forensics and
  Security (WIFS)}, pages 1--7. IEEE, 2018.

\bibitem{CelebDF}
Y.~Li, X.~Yang, P.~Sun, H.~Qi, and S.~Lyu.
\newblock Celeb-df: A large-scale challenging dataset for deepfake forensics.
\newblock In {\em Proceedings of the IEEE/CVF Conference on Computer Vision and
  Pattern Recognition}, pages 3207--3216, 2020.

\bibitem{TSM}
J.~Lin, C.~Gan, and S.~Han.
\newblock Tsm: Temporal shift module for efficient video understanding.
\newblock In {\em Proceedings of the IEEE/CVF International Conference on
  Computer Vision}, pages 7083--7093, 2019.

\bibitem{swin}
Z.~Liu, Y.~Lin, Y.~Cao, H.~Hu, Y.~Wei, Z.~Zhang, S.~Lin, and B.~Guo.
\newblock Swin transformer: Hierarchical vision transformer using shifted
  windows.
\newblock In {\em Proceedings of the IEEE/CVF International Conference on
  Computer Vision}, pages 10012--10022, 2021.

\bibitem{convNext}
Z.~Liu, H.~Mao, C.-Y. Wu, C.~Feichtenhofer, T.~Darrell, and S.~Xie.
\newblock A convnet for the 2020s.
\newblock {\em arXiv preprint arXiv:2201.03545}, 2022.

\bibitem{visual_artifacts}
F.~Matern, C.~Riess, and M.~Stamminger.
\newblock Exploiting visual artifacts to expose deepfakes and face
  manipulations.
\newblock In {\em 2019 IEEE Winter Applications of Computer Vision Workshops
  (WACVW)}, pages 83--92. IEEE, 2019.

\bibitem{DFGC2021}
B.~Peng, H.~Fan, W.~Wang, J.~Dong, Y.~Li, S.~Lyu, Q.~Li, Z.~Sun, H.~Chen,
  B.~Chen, et~al.
\newblock Dfgc 2021: A deepfake game competition.
\newblock In {\em 2021 IEEE International Joint Conference on Biometrics
  (IJCB)}, pages 1--8. IEEE, 2021.

\bibitem{unifiedSwap}
B.~Peng, H.~Fan, W.~Wang, J.~Dong, and S.~Lyu.
\newblock A unified framework for high fidelity face swap and expression
  reenactment.
\newblock {\em IEEE Transactions on Circuits and Systems for Video Technology},
  32(6):3673--3684, 2021.

\bibitem{DeepFaceLab}
I.~Perov, D.~Gao, N.~Chervoniy, K.~Liu, S.~Marangonda, C.~Um{\'e}, M.~Dpfks,
  C.~S. Facenheim, L.~RP, J.~Jiang, et~al.
\newblock Deepfacelab: Integrated, flexible and extensible face-swapping
  framework.
\newblock {\em arXiv preprint arXiv:2005.05535}, 2020.

\bibitem{FaceForensics++}
A.~Rössler, D.~Cozzolino, L.~Verdoliva, C.~Riess, J.~Thies, and M.~Niessner.
\newblock Faceforensics++: Learning to detect manipulated facial images.
\newblock In {\em 2019 IEEE/CVF International Conference on Computer Vision
  (ICCV)}, pages 1--11, 2019.

\bibitem{self_blend}
K.~Shiohara and T.~Yamasaki.
\newblock Detecting deepfakes with self-blended images.
\newblock In {\em Proceedings of the IEEE/CVF Conference on Computer Vision and
  Pattern Recognition}, pages 18720--18729, 2022.

\bibitem{meta_learn}
K.~Sun, H.~Liu, Q.~Ye, Y.~Gao, J.~Liu, L.~Shao, and R.~Ji.
\newblock Domain general face forgery detection by learning to weight.
\newblock In {\em Proceedings of the AAAI conference on artificial
  intelligence}, volume~35, pages 2638--2646, 2021.

\bibitem{efficientnet}
M.~Tan and Q.~Le.
\newblock Efficientnet: Rethinking model scaling for convolutional neural
  networks.
\newblock In {\em International conference on machine learning}, pages
  6105--6114. PMLR, 2019.

\bibitem{representative_mining}
C.~Wang and W.~Deng.
\newblock Representative forgery mining for fake face detection.
\newblock In {\em Proceedings of the IEEE/CVF conference on computer vision and
  pattern recognition}, pages 14923--14932, 2021.

\bibitem{Hififace}
Y.~Wang, X.~Chen, J.~Zhu, W.~Chu, Y.~Tai, C.~Wang, J.~Li, Y.~Wu, F.~Huang, and
  R.~Ji.
\newblock Hififace: 3d shape and semantic prior guided high fidelity face
  swapping.
\newblock In Z.-H. Zhou, editor, {\em Proceedings of the Thirtieth
  International Joint Conference on Artificial Intelligence, {IJCAI-21}}, pages
  1136--1142. International Joint Conferences on Artificial Intelligence
  Organization, 8 2021.
\newblock Main Track.

\bibitem{Multi-attention}
H.~Zhao, W.~Zhou, D.~Chen, T.~Wei, W.~Zhang, and N.~Yu.
\newblock Multi-attentional deepfake detection.
\newblock In {\em Proceedings of the IEEE/CVF conference on computer vision and
  pattern recognition}, pages 2185--2194, 2021.

\bibitem{patch_consistency}
T.~Zhao, X.~Xu, M.~Xu, H.~Ding, Y.~Xiong, and W.~Xia.
\newblock Learning self-consistency for deepfake detection.
\newblock In {\em Proceedings of the IEEE/CVF international conference on
  computer vision}, pages 15023--15033, 2021.

\bibitem{MegaFS}
Y.~Zhu, Q.~Li, J.~Wang, C.-Z. Xu, and Z.~Sun.
\newblock One shot face swapping on megapixels.
\newblock In {\em Proceedings of the IEEE/CVF Conference on Computer Vision and
  Pattern Recognition}, pages 4834--4844, 2021.

\bibitem{WildDF}
B.~Zi, M.~Chang, J.~Chen, X.~Ma, and Y.-G. Jiang.
\newblock Wilddeepfake: A challenging real-world dataset for deepfake
  detection.
\newblock In {\em Proceedings of the 28th ACM International Conference on
  Multimedia}, pages 2382--2390, 2020.

\end{thebibliography}
}

\end{document}